\documentclass[twoside,11pt]{article}

\usepackage{jmlr2e}
\usepackage{hyperref}

\begin{document}




\firstpageno{1}

\title{Recent Advances in Deep Learning: An Overview}

\author{\name Matiur Rahman Minar \email minar09.bd@gmail.com \\
       \name Jibon Naher \email jibon.naher09@gmail.com \\ 
       \addr Department of Computer Science and Engineering\\
       Chittagong University of Engineering and Technology\\
       Chittagong-4349, Bangladesh}

\editor{}

\maketitle


\begin{abstract}
Deep Learning is one of the newest trends in Machine Learning and Artificial Intelligence research. It is also one of the most popular scientific research trends now-a-days. Deep learning methods have brought revolutionary advances in computer vision and machine learning. Every now and then, new and new deep learning techniques are being born, outperforming state-of-the-art machine learning and even existing deep learning techniques. In recent years, the world has seen many major breakthroughs in this field. Since deep learning is evolving at a huge speed, its kind of hard to keep track of the regular advances especially for new researchers. In this paper, we are going to briefly discuss about recent advances in Deep Learning for past few years.
\end{abstract}



\section{Introduction}

The  term "Deep Learning" (DL) was first introduced to Machine Learning (ML) in 1986, and later used for Artificial Neural Networks (ANN) in 2000 \citep{Schmidhuber:2015}. Deep learning methods are composed of multiple layers to learn features of data with multiple levels of abstraction \citep{LeCun-et-al-2015}. DL approaches allow computers to learn complicated concepts by building them out of simpler ones \citep{Goodfellow-et-al-2016}. For Artificial Neural Networks (ANN), Deep Learning (DL) aka hierarchical learning \citep{Deng-et-al-2014} is about assigning credits in many computational stages accurately, to transform the aggregate activation of the network \citep{Schmidhuber-2014}. To learn complicated functions, deep architectures are used with multiple levels of abstractions i.e. non-linear operations; e.g. ANNs with many hidden layers \citep{Bengio-2009}. To sum it accurately, Deep Learning is a sub-field of Machine Learning, which uses many levels of non-linear information processing and abstraction, for supervised or unsupervised feature learning and representation, classification and pattern recognition \citep{Deng-et-al-2014}.

Deep Learning i.e. Representation Learning is class or sub-field of Machine Learning. Recent deep learning methods are mostly said to be developed since 2006 \citep{Deng:2011}. This paper is an overview of most recent techniques of deep learning, mainly recommended for upcoming researchers in this field. This article includes the basic idea of DL, major approaches and methods, recent breakthroughs and applications.

Overview papers are found to be very beneficial, especially for new researchers in a particular field. It is often hard to keep track with contemporary advances in a research area, provided that field has great value in near future and related applications. Now-a-days, scientific research is an attractive profession since knowledge and education are more shared and available than ever. For a technological research trend, its only normal to assume that there will be numerous advances and improvements in various ways. An overview of an particular field from couple years back, may turn out to be obsolete today.

Considering the popularity and expansion of Deep Learning in recent years, we present a brief overview of Deep Learning as well as Neural Networks (NN), and its major advances and critical breakthroughs from past few years. We hope that this paper will help many novice researchers in this field, getting an overall picture of recent Deep Learning researches and techniques, and guiding them to the right way to start with. Also we hope to pay some tributes by this work, to the top DL and ANN researchers of this era, Geoffrey Hinton \citep{Hinton-self}, Juergen Schmidhuber \citep{Schmidhuber-self}, Yann LeCun \citep{LeCun-self}, Yoshua Bengio \citep{Bengio-self} and many others who worked meticulously to shape the modern Artificial Intelligence (AI). Its also important to follow their works to stay updated with state-of-the-art in DL and ML research.

In this paper, firstly we will provide short descriptions of the past overview papers on deep learning models and approaches. Then, we will start describing the recent advances of this field. We are going to discuss Deep Learning (DL) approaches, deep architectures i.e. Deep Neural Networks (DNN) and Deep Generative Models (DGM), followed by important regularization and optimization methods. Also, there are two brief sections for open-source DL frameworks and significant DL applications. Finally, we will discuss about current status and the future of Deep Learning in the last two sections i.e. Discussion and Conclusion.

\section{Related works}
There were many overview papers on Deep Learning (DL) in the past years. They described DL methods and approaches in great ways as well as their applications and directions for future research. Here, we are going to brief some outstanding overview papers on deep learning.

\cite{DBLP:journals/corr/abs-1708-02709} talked about DL models and architectures, mainly used in Natural Language Processing (NLP). They showed DL applications in various NLP fields, compared DL models, and discussed possible future trends.

\cite{DBLP:journals/corr/ZhangGPMS17} discussed state-of-the-art deep learning techniques for front-end and back-end speech recognition systems.

\cite{DBLP:journals/corr/abs-1710-03959} presented overview on state-of-the-art of DL for remote sensing. They also discussed open-source DL frameworks and other technical details for deep learning.

\cite{DBLP:journals/corr/WangRX17} described the evolution of deep learning models in time-series manner. The briefed the models graphically along with the breakthroughs in DL research. This paper would be a good read to know the origin of the Deep Learning in evolutionary manner. They also mentioned optimization and future research of neural networks.

\cite{Goodfellow-et-al-2016} discussed deep networks and generative models in details. Starting from Machine Learning (ML) basics, pros and cons for deep architectures, they concluded recent DL researches and applications thoroughly. 

\cite{LeCun-et-al-2015} published a overview of Deep Learning (DL) models with Convolutional Neural Networks (CNN) and Recurrent Neural Networks (RNN). They described DL from the perspective of Representation Learning, showing how DL techniques work and getting used successfully in various applications, and predicting future learning based on Unsupervised Learning (UL). They also pointed out the articles of major advances in DL in the bibliography.

\cite{Schmidhuber:2015} did a generic and historical overview of Deep Learning along with CNN, RNN and Deep Reinforcement Learning (RL). He emphasized on sequence-processing RNNs, while pointing out the limitations of fundamental DL and NNs, and the tricks to improve them.

\cite{Nielsen-2015} described the neural networks in details along with codes and examples. He also discussed deep neural networks and deep learning to some extent.

\cite{Schmidhuber-2014} covered history and evolution of neural networks based on time progression, categorized with machine learning approaches, and uses of deep learning in the neural networks.

\cite{Deng-et-al-2014} described deep learning classes and techniques, and applications of DL in several areas.

\cite{DBLP:journals/corr/abs-1305-0445} did quick overview on DL algorithms i.e. supervised and unsupervised networks, optimization and training models from the perspective of representation learning. He focused on many challenges of Deep Learning e.g. scaling algorithms for larger models and data, reducing optimization difficulties, designing efficient scaling methods etc. along with optimistic DL researches.

\cite{Bengio:2013:RLR:2498740.2498889} discussed on Representation and Feature Learning aka Deep Learning. They explored various methods and models from the perspectives of applications, techniques and challenges. 

\cite{Deng:2011} gave an overview of deep structured learning and its architectures from the perspectives of information processing and related fields.

\cite{Arel:2010:RFD:1921914.1921920} provided a short overview on recent DL techniques.

\cite{Bengio-2009} discussed deep architectures i.e. neural networks and generative models for AI.

All recent overview papers on Deep Learning (DL) discussed important things from several perspectives. It is necessary to go through them for a DL researcher. However, DL is a highly flourishing field right now. Many new techniques and architectures are invented, even after the most recently published overview paper on DL. Also, previous papers focus from different perspectives. Our paper is mainly for the new learners and novice researchers who are new to this field. For that purpose, we will try to give a basic and clear idea of deep learning to the new researchers and anyone interested in this field.

\section{Recent Advances}
In this section, we will discuss the main recent Deep Learning (DL) approaches derived from Machine Learning and brief evolution of Artificial Neural Networks (ANN), which is the most common form used for deep learning.

\subsection{Evolution of Deep Architectures}
Artificial Neural Networks (ANN) have come a long way, as well as other deep models. First generation of ANNs was composed of simple neural layers for Perceptron. They were limited in simples computations. Second generation used Backpropagation to update weights of neurons according to error rates. Then Support Vector Machine (SVM) surfaced, and surpassed ANNs for a while. To overcome the limitations of backpropagation, Restricted Boltzmann Machine was proposed, making the learning easier. Other techniques and neural networks came as well e.g. Feedforward Neural Networks (FNN), Convolutional Neural Netowrks (CNN), Recurrent Neural Networks (RNN) etc. along with Deep Belief Networks, Autoencoders and such (\cite{Hinton-self}, \textit{The next generation of neural networks}). From that point, ANNs got improved and designed in various ways and for various purposes.

\cite{Schmidhuber-2014}, \cite{Bengio-2009}, \cite{Deng-et-al-2014}, \cite{Goodfellow-et-al-2016}, \cite{DBLP:journals/corr/WangRX17} etc. provided detailed overview on the evolution and history of Deep Neural Networks (DNN) as well as Deep Learning (DL). Deep architectures are multilayer non-linear repetition of simple architectures in most of the cases, which helps to obtain highly complex functions out of the inputs \citep{LeCun-et-al-2015}.

\section{Deep Learning Approaches}
Deep Neural Networks (DNN) gained huge success in Supervised Learning (SL). Also, Deep Learning (DL) models are immensely successful in Unsupervised, Hybrid and Reinforcement Learning as well \citep{LeCun-et-al-2015}.

\subsection{Deep Supervised Learning}
Supervised learning are applied when data is labeled and the classifier is used for class or numeric prediction. \cite{LeCun-et-al-2015} provided a brief yet very good explanation of supervised learning approach and how deep architectures are formed. \cite{Deng-et-al-2014} mentioned many deep networks for supervised and hybrid learning and explained them e.g. Deep Stacking Network (DSN) and its variants. \cite{Schmidhuber-2014} covered all neural networks starting from early neural networks to recently successful Convolutional Neural Networks (CNN), Recurrent Neural Networks (RNN), Long Short Term Memory (LSTM) and their improvements.

\subsection{Deep Unsupervised Learning}
When input data is not labeled, unsupervised learning approach is applied to extract features from data and classify or label them. \cite{LeCun-et-al-2015} predicted future of deep learning in unsupervised learning. \cite{Schmidhuber-2014} described neural networks for unsupervised learning as well. \cite{Deng-et-al-2014} briefed deep architectures for unsupervised learning and explained deep Autoencoders in detail.

\subsection{Deep Reinforcement Learning}
Reinforcement learning uses reward and punishment system for the next move generated by the learning model. This is mostly used for games and robots, solves usually decision making problems \citep{DBLP:journals/corr/Li17b}.
\cite{Schmidhuber-2014} described advances of deep learning in Reinforcement Learning (RL) and uses of Deep Feedforward Neural Netowrk (FNN) and Recurrent Neural Network (RNN) for RL. \cite{DBLP:journals/corr/Li17b} discussed Deep Reinforcement Learning(DRL), its architectures e.g. Deep Q-Network (DQN), and applications in various fields.

\cite{DBLP:journals/corr/MnihBMGLHSK16} proposed a DRL framework using asynchronous gradient descent for DNN optimization.

\cite{DBLP:journals/corr/HasseltGS15} proposed a DRL architecture using deep neural network (DNN).

\section{Deep Neural Networks}
In this section, we will briefly discuss about the deep neural networks (DNN), and recent improvements and breakthroughs of them. Neural networks work with functionalities similar to human brain. These are composed on neurons and connections mainly. When we are saying deep neural network, we can assume there should be quite a number of hidden layers, which can be used to extract features from the inputs and to compute complex functions. \cite{Bengio-2009} explained neural networks for deep architectures e.g. Convolutional Neural Networks (CNN), Auto-Encoders (AE) etc. and their variants. \cite{Deng-et-al-2014} detailed some neural network architectures e.g. AE and its variants. \cite{Goodfellow-et-al-2016} wrote and skillfully explained about Deep Feedforward Networks, Convolutional Networks, Recurrent and Recursive Networks and their improvements. \cite{Schmidhuber-2014} mentioned full history of neural networks from early neural networks to recent successful techniques.

\subsection{Deep Autoencoders}
Autoencoders (AE) are neural networks (NN) where outputs are the inputs. AE takes the original input, encodes for compressed representation and then decodes to reconstruct the input \citep{Wang-self}. In a deep AE, lower hidden layers are used for encoding and higher ones for decoding, and error back-propagation is used for training \citep{Deng-et-al-2014}. \cite{Goodfellow-et-al-2016}

\subsubsection{Variational Autoencoders}
Variational Auto-Encoders (VAE) can be counted as decoders \citep{Wang-self}. VAEs are built upon standard neural networks and can be trained with stochastic gradient descent \citep{Doersch2016}

\subsubsection{Stacked Denoising Autoencoders}
In early Auto-Encoders (AE), encoding layer had smaller dimensions than the input layer. In Stacked Denoising Auto-Encoders (SDAE), encoding layer is wider than the input layer \citep{Deng-et-al-2014}.

\subsubsection{Transforming Autoencoders}
Deep Auto-Encoders (DAE) can be transformation-variant, i.e., the extracted features from multilayers of non-linear processing could be changed due to learner. Transforming Auto-Encoders (TAE) work with both input vector and target output vector to apply transformation-invariant property and lead the codes towards a desired way \citep{Deng-et-al-2014}.

\subsection{Deep Convolutional Neural Networks}
Four basic ideas make the Convolutional Neural Networks (CNN), i.e., local connections, shared weights, pooling, and using many layers. First parts of a CNN are made of convolutional and pooling layers and latter parts are mainly fully connected layers. Convolutional layers detect local conjunctions from features and pooling layers merge similar features into one \citep{LeCun-et-al-2015}. CNNs use convolutions instead of matrix multiplication in the convolutional layers \citep{Goodfellow-et-al-2016}.

\cite{Krizhevsky:2012:ICD:2999134.2999257} presented a Deep Convolutional Neural Network (CNN) architecture, also known as AlexNet, which was a major breakthrough in Deep Learning (DL). The network composed of five convolutional layers and three fully connected layers. The architecture used Graphics Processing Units (GPU) for convolution operation, Rectified Linear Units (ReLU) as activation function and Dropout \citep{JMLR:v15:srivastava14a} to reduce overfitting.

\cite{DBLP:journals/corr/IandolaMAHDK16} proposed a small CNN architecture called SqueezeNet.

\cite{DBLP:journals/corr/SzegedyLJSRAEVR14} proposed a Deep CNN architecture named Inception. An improvement of Inception-ResNet is proposed by \cite{DBLP:journals/corr/DaiQXLZHW17}.

\cite{DBLP:journals/corr/RedmonDGF15} proposed a CNN architecture named YOLO (You Only Look Once) for unified and real-time object detection.

\cite{DBLP:journals/corr/ZeilerF13} proposed a method for visualizing the activities within CNN.

\cite{DBLP:journals/corr/GehringAGYD17} proposed a CNN architecture for sequence-to-sequence learning.

\cite{DBLP:journals/corr/BansalCRGR17} proposed PixelNet, using pixels for representations.

\cite{Goodfellow-et-al-2016} explained the basic CNN architecures and the ideas. \cite{DBLP:journals/corr/GuWKMSSLWW15} presented a nice overview on recent advances of CNNs, multiple variants of CNN, its architectures, regularization methods and functionality, and applications in various fields.

\subsubsection{Deep Max-Pooling Convolutional Neural Networks}
Max-Pooling Convolutional Neural Networks (MPCNN) operate on mainly convolutions and max-pooling, especially used in digital image processing. MPCNN generally consists of three types of layers other than the input layer. Convolutional layers take input images and generate maps, then apply non-linear activation function. Max-pooling layers down-sample images and keep the maximum value of a sub-region. And fully-connected layers does the linear multiplication \citep{DBLP:conf/icip/MasciGCFS13}. In Deep MPCNN, convolutional and max-pooling layers are used periodically after the input layer, followed by fully-connected layers \citep{DBLP:conf/icip/GiustiCMGS13}.

\subsubsection{Very Deep Convolutional Neural Networks}
\cite{DBLP:journals/corr/SimonyanZ14a} proposed Very Deep Convolutional Neural Network (VDCNN) architecture, also known as VGG Nets. VGG Nets use very small convolution filters and depth to 16-19 weight layers.

\cite{DBLP:journals/corr/ConneauSBL16} proposed another VDCNN architecture for text classification which uses small convolutions and pooling. They claimed this architecture is the first VDCNN to be used in text processing which works at the character level. This architecture is composed of 29 convolution layers.

\subsection{Network In Network}
\cite{DBLP:journals/corr/LinCY13} proposed Network In Network (NIN). NIN replaces convolution layers of traditional Convolutional Neural Network (CNN) by micro neural networks with complex structures. It uses multi-layer perceptron (MLPConv) for micro neural networks and global average pooling layer instead of fully connected layers. Deep NIN architectures can be made from multi-stacking of this proposed NIN structure \citep{DBLP:journals/corr/LinCY13}.

\subsection{Region-based Convolutional Neural Networks}
\cite{girshick2014rcnn} proposed Region-based Convolutional Neural Network (R-CNN) which uses regions for recognition. R-CNN uses regions to localize and segment objects. This architecture consists of three modules i.e. category independent region proposals which defines the set of candidate regions, large Convolutional Neural Network (CNN) for extracting features from the regions, and a set of class specific linear Support Vector Machines (SVM) \citep{girshick2014rcnn}.

\subsubsection{Fast R-CNN}
\cite{DBLP:journals/corr/Girshick15} proposed Fast Region-based Convolutional Network (Fast R-CNN). This method exploits R-CNN \citep{girshick2014rcnn} architecture and produces fast results. Fast R-CNN consists of convolutional and pooling layers, proposals of regions, and a sequence of fully connected layers \citep{DBLP:journals/corr/Girshick15}.

\subsubsection{Faster R-CNN}
\cite{ren2015faster} proposed Faster Region-based Convolutional Neural Networks (Faster R-CNN), which uses Region Proposal Network (RPN) for real-time object detection. RPN is a fully convolutional network which generates region proposals accurately and efficiently \citep{ren2015faster}.

\subsubsection{Mask R-CNN}
\cite{DBLP:journals/corr/HeGDG17} proposed Mask Region-based Convolutional Network (Mask R-CNN) instance object segmentation. Mask R-CNN extends Faster R-CNN \citep{ren2015faster} architecture, and uses an extra branch for object mask \citep{DBLP:journals/corr/HeGDG17}.

\subsubsection{Multi-Expert R-CNN}
\cite{DBLP:journals/corr/LeeEK17} proposed Multi-Expert Region-based Convolutional Neural Networks (ME R-CNN), which exploits Fast R-CNN \citep{DBLP:journals/corr/Girshick15} architecture. ME R-CNN generates Region of Interests (RoI) from selective and exhaustive search. Also it uses per-RoI multi-expert network instead of single per-RoI network. Each expert is the same architecture of fully connected layers from Fast R-CNN \citep{DBLP:journals/corr/LeeEK17}.

\subsection{Deep Residual Networks}
\cite{DBLP:journals/corr/HeZRS15} proposed Residual Networks (ResNets) consists of 152 layers. ResNets have lower error and easily trained with Residual Learning. More deeper ResNets achieve more better performance \citep{He-self}. ResNets are considered an important advance in the field of Deep Learning.

\subsubsection{Resnet in Resnet}
\cite{DBLP:journals/corr/TargAL16} proposed Resnet in Resnet (RiR) which combines ResNets \citep{DBLP:journals/corr/HeZRS15} and standard Convolutional Neural Networks (CNN) in a deep dual stream architecture \citep{DBLP:journals/corr/TargAL16}.

\subsubsection{ResNeXt}
\cite{DBLP:journals/corr/XieGDTH16} proposed ResNeXt architecture. ResNext  exploits ResNets \citep{DBLP:journals/corr/HeZRS15} for repeating layers with split-transform-merge strategy \citep{DBLP:journals/corr/XieGDTH16}.

\subsection{Capsule Networks}
\cite{DBLP:conf/nips/SabourFH17} proposed Capsule Networks (CapsNet), an architecture with two convolutional layers and one fully connected layer. CapsNet usually contains several convolution layers and on capsule layer at the end \citep{Xi-et-al-2017}. CapsNet is considered as one of the most recent breakthrough in Deep Learning \citep{Xi-et-al-2017}, since this is said to be build upon the limitations of Convolutional Neural Networks \citep{Hinton-self}. It uses layers of capsules instead of layers of neurons, where a capsule is a set of neurons. Active lower level capsules make predictions and upon agreeing multiple predictions, a higher level capsule becomes active. A routing-by-agreement mechanism is used in these capsule layers. An improvement of CapsNet is proposed with EM routing \citep{anonymous2018matrix} using Expectation-Maximization (EM) algorithm.

\subsection{Recurrent Neural Networks}
Recurrent Neural Networks (RNN) are better suited for sequential inputs like speech and text and generating sequence. A Recurrent hidden unit can be considered as very deep feedforward network with same weights when unfolded in time. RNNs used to be difficult to train because of gradient vanishing and exploding problem \citep{LeCun-et-al-2015}. Many improvements were proposed later to solve this problem.

\cite{Goodfellow-et-al-2016} provided details of Recurrent and Recursive Neural Networks and architectures, its variants along with related gated and memory networks.

\cite{DBLP:journals/corr/KarpathyJL15} used character-level language models for analyzing and visualizing predictions, representations training dynamics, and error types of RNN and its variants e.g. LSTMs.

\cite{DBLP:journals/corr/JozefowiczVSSW16} explored RNN models and limitations for language modelling.

\subsubsection{RNN-EM}
\cite{DBLP:conf/nlpcc/PengYJW15} proposed Recurrent Neural Networks with External Memory (RNN-EM) to improve memory capacity of RNNs. They claimed to achieve state-of-the-art in language understanding, better than other RNNs.

\subsubsection{GF-RNN}
\cite{Chung:2015:GFR:3045118.3045338} proposed Gated Feedback Recurrent Neural Networks (GF-RNN), which extends the standard RNN by stacking multiple recurrent layers with global gating units.

\subsubsection{CRF-RNN}
\cite{DBLP:journals/corr/ZhengJRVSDHT15} proposed Conditional Random Fields as Recurrent Neural Networks (CRF-RNN), which combines the Convolutional Neural Networks (CNNs) and Conditional Random Fields (CRFs) for probabilistic graphical modelling.

\subsubsection{Quasi-RNN}
\cite{DBLP:journals/corr/BradburyMXS16} proposed Quasi Recurrent Neural Networks (QRNN) for neural sequence modelling, appling parallel across timesteps.

\subsection{Memory Networks}
\cite{DBLP:journals/corr/WestonCB14} proposed Memory Networks for question answering (QA). Memory Networks are composed of memory, input feature map, generalization, output feature map and response \citep{DBLP:journals/corr/WestonCB14} .

\subsubsection{Dynamic Memory Networks}
\cite{DBLP:journals/corr/KumarISBEPOGS15} proposed Dynamic Memory Networks (DMN) for QA tasks. DMN has four modules i.e. Input, Question, Episodic Memory, Output \citep{DBLP:journals/corr/KumarISBEPOGS15}.

\subsection{Augmented Neural Networks}
\cite{olah2016attention} gave nice presentation of Attentional and Augmented Recurrent Neural Networks i.e. Neural Turing Machines (NTM), Attentional Interfaces, Neural Programmer and Adaptive Computation Time. Augmented Neural Networks are usually made of using extra properties like logic functions along with standard Neural Network architecture \citep{olah2016attention}.

\subsubsection{Neural Turing Machines}
\cite{DBLP:journals/corr/GravesWD14} proposed Neural Turing Machine (NTM) architecture, consisting of a neural network controller and a memory bank. NTMs usually combine RNNs with external memory bank \citep{olah2016attention}.

\subsubsection{Neural GPU}
\cite{DBLP:journals/corr/KaiserS15} proposed Neural GPU, which solves the parallel problem of NTM \citep{DBLP:journals/corr/GravesWD14}.

\subsubsection{Neural Random-Access Machines}
\cite{DBLP:journals/corr/KurachAS15} proposed Neural Random Access Machine, which uses an external variable-size random-access memory.

\subsubsection{Neural Programmer}
\cite{DBLP:journals/corr/NeelakantanLS15} proposed Neural Programmer, an augmented neural network with arithmetic and logic functions.

\subsubsection{Neural Programmer-Interpreters}
\cite{DBLP:journals/corr/ReedF15} proposed Neural Programmer-Interpreters (NPI) which can learn. NPI consists of recurrent core, program memory and domain-specific encoders \citep{DBLP:journals/corr/ReedF15}.

\subsection{Long Short Term Memory Networks}
\cite{Hochreiter:1997:LSM:1246443.1246450} proposed Long Short-Term Memory (LSTM) which overcomes the error back-flow problems of Recurrent Neural Networks (RNN). LSTM is based on recurrent network along with gradient-based learning algorithm \citep{Hochreiter:1997:LSM:1246443.1246450} LSTM introduced self-loops to produce paths so that gradient can flow \citep{Goodfellow-et-al-2016}.

\cite{DBLP:journals/tnn/GreffSKSS17} provided large-scale analysis of Vanilla LSTM and eight LSTM variants for three uses i.e. speech recognition, handwriting recognition, and polyphonic music modeling. They claimed that eight variants of LSTM failed to perform significant improvement, while only Vanilla LSTM performs well \citep{DBLP:journals/corr/GreffSKSS15}.

\cite{DBLP:conf/naacl/ShiYTJ16} proposed Deep Long Short-Term Memory (DLSTM), which is a stack of LSTM units for feature mapping to learn representations \citep{DBLP:conf/naacl/ShiYTJ16}.

\subsubsection{Batch-Normalized LSTM}
\cite{DBLP:journals/corr/CooijmansBLC16} proposed batch-normalized LSTM (BN-LSTM), which uses batch-normalizing on hidden states of recurrent neural networks.

\subsubsection{Pixel RNN}
\cite{DBLP:journals/corr/OordKK16} proposed Pixel Recurrent Neural Networks (PixelRNN), made of up to twelve two-dimensional LSTM layers.

\subsubsection{Bidirectional LSTM}
\cite{WÃ¶llmer2010} proposed Bidirection LSTM (BLSTM) Recurrent Networks to be used with Dynamic Bayesian Network (DBN) for context-sensitive keyword detection.

\subsubsection{Variational Bi-LSTM}
\cite{Shabanian2017} proposed Variational Bi-LSTMs, which is a variant of Bidirectional LSTM architecture. Variational Bi-LSTM creates a channel of information exchange between LSTMs using Variational Auto-Encoders (VAE), for learning better representations \citep{Shabanian2017}.

\subsection{Google’s Neural Machine Translation}
\cite{DBLP:journals/corr/WuSCLNMKCGMKSJL16} proposed Google’s Neural Machine Translation (GNMT) System for automated translation, which incorporates an encoder network, a decoder network and an attention network following the common sequence-to-sequence learning framework.

\subsection{Fader Networks}
\cite{DBLP:journals/corr/LampleZUBDR17} proposed Fader Networks, a new type of encoder-decoder architecture to generate realistic variations of input images by changing attribute values.

\subsection{Hyper Networks}
\cite{DBLP:journals/corr/HaDL16} proposed HyperNetworks which generates weights for other neural networks, such as static hypernetworks convolutional networks, dynamic hypernetworks for recurrent networks.

\cite{Deutsch-2018} used Hyper Networks for generating neural networks.

\subsection{Highway Networks}
\cite{DBLP:journals/corr/SrivastavaGS15} proposed Highway Networks, which uses gating units to learn regulating information through. Information flow across several layers are called information highways \citep{DBLP:journals/corr/SrivastavaGS15}.

\subsubsection{Recurrent Highway Networks}
\cite{DBLP:conf/icml/ZillySKS17} proposed Recurrent Highway Networks (RHN), which extend Long Short-Term Memory (LSTM) architecture. RHNs use Highway layers inside the recurrent transition \citep{DBLP:conf/icml/ZillySKS17}.

\subsection{Highway LSTM RNN}
\cite{DBLP:conf/icassp/ZhangCYYKG16} proposed Highway Long Short-Term Memory (HLSTM) RNN, which extends deep LSTM networks with gated direction connections i.e. Highways, between memory cells in adjacent layers.

\subsection{Long-Term Recurrent CNN}
\cite{DBLP:journals/corr/DonahueHGRVSD14} proposed Long-term Recurrent Convolutional Networks (LRCN), which uses CNN for inputs, then LSTM for recurrent sequence modeling and generating predictions.

\subsection{Deep Neural SVM}
\cite{DBLP:conf/icassp/ZhangLYG15} proposed Deep Neural Support Vector Machines (DNSVM), which uses Support Vector Machine (SVM) as the top layer for classification in a Deep Neural Network (DNN)

\subsection{Convolutional Residual Memory Networks}
\cite{DBLP:journals/corr/MonizP16} proposed Convolutional Residual Memory Networks, which incorporates memory mechanism into Convolutional Neural Networks (CNN). It augments convolutional residual networks with a long short term memory mechanism \citep{DBLP:journals/corr/MonizP16}.

\subsection{Fractal Networks}
\cite{DBLP:journals/corr/LarssonMS16a} proposed Fractal Networks i.e. FractalNet, as an alternative to residual nets. They claimed to train ultra deep neural networks without residual learning. Fractals are repeated architecture generated by simple expansion rule \citep{DBLP:journals/corr/LarssonMS16a}.

\subsection{WaveNet}
\cite{DBLP:journals/corr/OordDZSVGKSK16} proposed WaveNet, deep neural network for generating raw audio. WaveNet is composed of a stack of convolutional layers, and softmax distribution layer for outputs \citep{DBLP:journals/corr/OordDZSVGKSK16}.

\cite{DBLP:journals/corr/RethagePS17} proposed a WaveNet model for speech denoising.

\subsection{Pointer Networks}
\cite{Vinyals2017} proposed Pointer Networks (Ptr-Nets), which solves the problem of representing variable dictionaries by using a softmax probability distribution called "Pointer".

\section{Deep Generative Models}
In this section, we will briefly discuss other deep architecures which uses multiple levels of abstraction and representation similar to deep neural networks, also known as Deep Generative Models (DGM). \cite{Bengio-2009} explained deep architectures e.g. Boltzmann Machines (BM) and Restricted Boltzmann Machines (RBM) etc. and their variants.

\cite{Goodfellow-et-al-2016} explained deep generative models in details e.g. Restricted and Unrestricted Boltzmann Machines and their variants, Deep Boltzmann Machines, Deep Belief Networks (DBN), Directed Generative Nets, and Generative Stochastic Networks etc.

\cite{Maaloe:2016:ADG:3045390.3045543} proposed Auxiliary Deep Generative Models where they extended Deep Generative Models with auxiliary variables. The auxiliary variables make variational distribution with stochastic layers and skip connections \citep{Maaloe:2016:ADG:3045390.3045543}.

\cite{pmlr-v48-rezende16} developed a class for one-shot generalization of deep generative models.

\subsection{Boltzmann Machines}
Boltzmann Machines are connectionist approach for learning arbitrary probability distributions which use maximum likelihood principle for learning  \citep{Goodfellow-et-al-2016}.

\subsection{Restricted Boltzmann Machines}
Restricted Boltzmann Machines (RBM) are special type of Markov random field containing one layer of stochastic hidden units i.e. latent variables and one layer of observable variables (\cite{Deng-et-al-2014}, \cite{Goodfellow-et-al-2016}).

\cite{TOPS:TOPS1109} proposed a Deep Generative Model using Restricted Boltzmann Machines (RBM) for document processing.

\subsection{Deep Belief Networks}
Deep Belief Networks (DBN) are generative models with several layers of latent binary or real variables \citep{Goodfellow-et-al-2016}.

\cite{Ranzato:2011:DGM:2191740.2191850} built a deep generative model using Deep Belief Network (DBN) for images recognition.

\subsection{Deep Lambertian Networks}
\cite{Tang-et-al-2012} proposed Deep Lambertian Networks (DLN) which is a multilayer generative model where latent variables are albedo, surface normals, and the light source. DLN is a combination of lambertian reflectance with Gaussian Restricted Boltzmann Machines and Deep Belief Networks \citep{Tang-et-al-2012}.

\subsection{Generative Adversarial Networks}
\cite{NIPS2014_5423} proposed Generative Adversarial Nets (GAN) for estimating generative models with an adversarial process. GAN architecture is composed of a generative model pitted against an adversary i.e. a discriminative model to learn model or data distribution \citep{NIPS2014_5423}. Some more improvements proposed for GAN by \cite{DBLP:journals/corr/MaoLXLW16}, \cite{DBLP:journals/corr/KimCKLK17} etc.

\cite{DBLP:journals/corr/SalimansGZCRC16} presented several methods for training GANs.

\subsubsection{Laplacian Generative Adversarial Networks}
\cite{DBLP:journals/corr/DentonCSF15} proposed a Deep Generative Model (DGM) called Laplacian Generative Adversarial Networks (LAPGAN) using Generative Adversarial Networks (GAN) approach. The model also uses convolutional networks within a Laplacian pyramid framework \citep{DBLP:journals/corr/DentonCSF15}.

\subsection{Recurrent Support Vector Machines}
\cite{DBLP:conf/naacl/ShiYCYPH16} proposed Recurrent Support Vector Machines (RSVM), which uses Recurrent Neural Network (RNN) for extracting features from input sequence and standard Support Vector Machine (SVM) for sequence-level objective discrimination.

\section{Training and Optimization Techniques}
In this section, we will provide short overview on some major techniques for regularization and optimization of Deep Neural Networks (DNN).

\subsection{Dropout}
\cite{JMLR:v15:srivastava14a} proposed Dropout to prevent neural networks from overfitting. Dropout is a neural network model-averaging regularization method by adding noise to its hidden units. It drops units from the neural network along with connections randomly during training. Dropout can be used with any kind of neural networks, even in graphical models like RBM \citep{JMLR:v15:srivastava14a}. A very recent proposed improvement of dropout is Fraternal Dropout \citep{anonymous2018fraternal} for Recurrent Neural Networks (RNN).

\subsection{Maxout}
\cite{pmlr-v28-goodfellow13} proposed Maxout, a new activation function to be used with Dropout \citep{JMLR:v15:srivastava14a}. Maxout's output is the maximum of a set of inputs, which is beneficial for Dropout's model averaging \citep{pmlr-v28-goodfellow13}.

\subsection{Zoneout}
\cite{DBLP:journals/corr/KruegerMKPBKGBL16} proposed Zoneout, a regularization method for Recurrent Neural Networks (RNN). Zoneout uses noise randomly while training similar to Dropout \citep{JMLR:v15:srivastava14a}, but preserves hidden units instead of dropping \citep{DBLP:journals/corr/KruegerMKPBKGBL16}.

\subsection{Deep Residual Learning}
\cite{DBLP:journals/corr/HeZRS15} proposed Deep Residual Learning framework for Deep Neural Networks (DNN), which are called ResNets with lower training error \citep{He-self}.

\subsection{Batch Normalization}
\cite{pmlr-v37-ioffe15} proposed Batch Normalization, a method for accelerating deep neural network training by reducing internal covariate shift.
\cite{DBLP:journals/corr/Ioffe17} proposed Batch Renormalization extending the previous approach.

\subsection{Distillation}
\cite{44873} proposed Distillation, from transferring knowledge from ensemble of highly regularized models i.e. neural networks into compressed and smaller model.

\subsection{Layer Normalization}
\cite{DBLP:journals/corr/BaKH16} proposed Layer Normalization, for speeding-up training of deep neural networks especially for RNNs and solves the limitations of batch normalization \citep{pmlr-v37-ioffe15}.

\section{Deep Learning frameworks}
There are a good number of open-source libraries and frameworks available for deep learning. Most of them are built for python programming language. Such as Theano \citep{Bergstra2011TheanoAC}, Tensorflow \citep{DBLP:journals/corr/AbadiABBCCCDDDG16}, PyTorch, PyBrain \citep{Schaul:2010:PYB:1756006.1756030}, Caffe \citep{DBLP:journals/corr/JiaSDKLGGD14}, Blocks and Fuel \citep{DBLP:journals/corr/MerrienboerBDSW15}, CuDNN \citep{DBLP:journals/corr/ChetlurWVCTCS14}, Honk \citep{DBLP:journals/corr/abs-1710-06554}, ChainerCV \citep{DBLP:journals/corr/abs-1708-08169}, PyLearn2, Chainer, torch, neon etc. 

\cite{DBLP:journals/corr/BahrampourRSS15} did a comparative study of several deep learning frameworks.

\section{Applications of Deep Learning}
In this section, we will briefly discuss some recent outstanding applications of Deep Learning architectures. Since the beginning of Deep Learning (DL), DL methods are being used in various fields in forms of supervised, unsupervised, semi-supervised or reinforcement learning. Starting from classification and detection tasks, DL applications are spreading rapidly in every fields.

Such as - 
\begin{itemize}

\item image classification and recognition (\cite{DBLP:journals/corr/SimonyanZ14a}, \cite{Krizhevsky:2012:ICD:2999134.2999257}, \cite{DBLP:journals/corr/HeZRS15})

\item video classification \citep{Karpathy:2014:LVC:2679600.2680211}

\item sequence generation \citep{DBLP:journals/corr/Graves13}
\item defect classification \citep{DBLP:conf/ijcnn/MasciMFS13}

\item text, speech, image and video processing \citep{LeCun-et-al-2015}
\item text classification \citep{DBLP:journals/corr/ConneauSBL16}

\item speech processing \citep{Arel09adeep}

\item speech recognition and spoken language understanding (\cite{38131}, \cite{DBLP:journals/corr/ZhangCYYKG15}, \cite{DBLP:conf/icassp/ZhangCYYKG16}, \cite{DBLP:conf/icassp/ZhangCYYKG16}, \cite{DBLP:conf/icassp/ZhangLYG15}, \cite{DBLP:conf/naacl/ShiYCYPH16}, \cite{DBLP:journals/taslp/MesnilDYBDHHHTY15}, \cite{DBLP:conf/nlpcc/PengYJW15}, \cite{DBLP:journals/corr/AmodeiABCCCCCCD15})

\item text-to-speech generation (\cite{DBLP:journals/corr/WangSSWWJYXCBLA17}, \cite{DBLP:journals/corr/ArikCCDGKLMRSS17})

\item query classification \citep{DBLP:conf/naacl/ShiYTJ16}
\item sentence classification \citep{DBLP:journals/corr/Kim14f}
\item sentence modelling \citep{DBLP:journals/corr/KalchbrennerGB14}

\item word processing \citep{DBLP:journals/corr/abs-1301-3781}
\item premise selection \citep{DBLP:journals/corr/AlemiCISU16}

\item document and sentence processing (\cite{DBLP:journals/corr/LeM14}, \cite{DBLP:journals/corr/MikolovSCCD13})

\item generating image captions (\cite{DBLP:journals/corr/VinyalsTBE14}, \cite{DBLP:journals/corr/XuBKCCSZB15})

\item photographic style transfer \citep{DBLP:journals/corr/LuanPSB17}

\item natural image manifold \citep{DBLP:journals/corr/ZhuKSE16}
\item image colorization \citep{DBLP:journals/corr/ZhangIE16}
\item image question answering \citep{DBLP:journals/corr/YangHGDS15}

\item generating textures and stylized images \citep{DBLP:journals/corr/UlyanovLVL16}

\item visual and textual question answering (\cite{DBLP:journals/corr/XiongMS16}, \cite(DBLP:journals/corr/AntolALMBZP15))

\item visual recognition and description (\cite{DBLP:journals/corr/DonahueHGRVSD14}, \cite{DBLP:journals/corr/RazavianASC14}, \cite{Oquab:2014:LTM:2679600.2680210})

\item object detection (\cite{DBLP:journals/corr/LeeEK17}, \cite{Ranzato:2011:DGM:2191740.2191850}, \cite{DBLP:journals/corr/RedmonDGF15}, \cite{DBLP:journals/corr/LiuAESR15})

\item document processing \citep{TOPS:TOPS1109}

\item character motion synthesis and editing \citep{Holden:2016:DLF:2897824.2925975}

\item singing synthesis \citep{DBLP:journals/corr/BlaauwB17}

\item person identification \citep{li2014deepreid}
\item face recognition and verification \citep{Taigman:2014:DCG:2679600.2680208}
\item action recognition in videos \citep{DBLP:journals/corr/SimonyanZ14}
\item human action recognition \citep{Ji:2013:CNN:2412386.2412939}
\item action recognition \citep{DBLP:journals/corr/SharmaKS15}

\item classifying and visualizing motion capture sequences \citep{DBLP:journals/corr/ChoC13}

\item handwriting generation and prediction \citep{carter2016experiments}

\item automated and machine translation (\cite{DBLP:journals/corr/WuSCLNMKCGMKSJL16}, \cite{DBLP:journals/corr/ChoMGBSB14}, \cite{DBLP:journals/corr/BahdanauCB14}, \cite{DBLP:journals/corr/HermannKGEKSB15}, \cite{DBLP:journals/corr/LuongPM15})

\item named entity recognition \citep{DBLP:journals/corr/LampleBSKD16}

\item mobile vision \citep{DBLP:journals/corr/HowardZCKWWAA17}
\item conversational agents \citep{DBLP:journals/corr/GhazvininejadBC17}
\item calling genetic variants \citep{46409}
\item cancer detection \citep{Cruz-Roa2013}
\item X-ray CT reconstruction \citep{DBLP:journals/corr/KanMY16}
\item Epileptic Seizure Prediction \citep{mirowski-mlsp-08}
\item hardware acceleration \citep{DBLP:journals/corr/HanLMPPHD16}
\item robotics \citep{DBLP:journals/corr/abs-1301-3592}
\end{itemize}
to name a few.

\cite{Deng-et-al-2014} provided detailed lists of DL applications in various categories e.g. speech and audio processing, information retrieval, object recognition and computer vision, multimodal and multi-task learning etc.

Using Deep Reinforcement Learning (DRL) for mastering games has become a hot topic now-a-days. Every now and then, AI bots created with DNN and DRL, are beating human world champions and grandmasters in strategical and other games, from only hours of training. For example, AlphaGo and AlphaGo Zero for game of GO (\cite{Silver-et-al-GO-2017}, \cite{Silver-et-al-GO-NN-tree-2017}, \cite{DBLP:journals/corr/abs-1711-09091}), Dota2 (\cite{Batsford-2014}), Atari (\cite{DBLP:journals/corr/MnihKSGAWR13},\cite{Mnih-et-al-2015}, \cite{DBLP:journals/corr/HasseltGS15}), Chess and Shougi \citep{Silveretal2017}.

\section{Discussion}
Though Deep Learning has achieved tremendous success in many areas, it still has long way to go. There are many rooms left for improvement. As for limitations, the list is quite long as well. For example, \cite{DBLP:journals/corr/NguyenYC14} showed that Deep Neural Networks (DNN) can be easily fooled while recognizing images. There are other issues like transferability of features learned \citep{Yosinski:2014:TFD:2969033.2969197}. \cite{DBLP:journals/corr/HuangPGDA17} proposed an architecture for adersarial attacks on neural networks, where they think future works are needed for defenses against those attacks. \cite{DBLP:journals/corr/ZhangBHRV16} presented an experimental framework for understanding deep learning models. They think understanding deep learning requires rethinking generalization.

\cite{Marcus2018} gave an important review on Deep Learning (DL), what it does, its limits and its nature. He strongly pointed out the limitations of DL methods, i.e., requiring more data, having limited capacity, inability to deal with hierarchical structure, struggling with open-ended inference, not being sufficiently transparent, not being well integrated with prior knowledge, and inability to distinguish causation from correlation \citep{Marcus2018}. He also mentioned that DL assumes stable world, works as approximation, is difficult to engineer and has potential risks as being an excessive hype. \cite{Marcus2018} thinks DL needs to be reconceptualized and to look for possibilities in unsupervised learning, symbol manipulation and hybrid models, having insights from cognitive science and psychology and taking bolder challenges.

\section{Conclusion}
Although Deep Learning (DL) has advanced the world faster than ever, there are still ways to go. We are still away from fully understanding of how deep learning works, how we can get machines more smarter, close to or smarter than humans, or learning exactly like human. DL has been solving many problems while taking technologies to another dimension. However, there are many difficult problems for humanity to deal with. For example, people are still dying from hunger and food crisis, cancer and other lethal diseases etc. We hope deep learning and AI will be much more devoted to the betterment of humanity, to carry out the hardest scientific researches, and last but not the least, to make the world a more better place for every single human.








\acks{We would like to thank Dr. Mohammed Moshiul Hoque, Professor, Department of CSE, CUET, for introducing us to the amazing world of Deep Learning.}

\bibliography{recent-deep-learning.bib}








\end{document}